\pdfoutput=1

\documentclass[11pt]{article}

\usepackage[]{acl}
\usepackage{times}
\usepackage{latexsym}
\usepackage{multirow}
\usepackage[T1]{fontenc}
\usepackage{booktabs}
\usepackage{bbding}
\usepackage{amsfonts,amssymb}

\usepackage[utf8]{inputenc}
\usepackage{subfigure}
\usepackage{parskip}
\usepackage{diagbox}
\usepackage{graphicx}
\usepackage{float}
\usepackage{caption}
\usepackage{amsmath}
\usepackage{microtype}

\title{Expose Backdoors on the Way: A Feature-Based Efficient Defense \\against Textual Backdoor Attacks}

\author{Sishuo Chen\textsuperscript{1}, Wenkai Yang\textsuperscript{1}, Zhiyuan Zhang\textsuperscript{2}, Xiaohan Bi\textsuperscript{1}, Xu Sun\textsuperscript{2} \\
  \textsuperscript{1}Center for Data Science, Peking University\\
  \textsuperscript{2}MOE Key Laboratory of Computational Linguistics, School of Computer Science, \\ Peking University\\
    \texttt{\{chensishuo,zzy1210,xusun\}@pku.edu.cn} \\
  \texttt{\{wkyang,bxh\}@stu.pku.edu.cn} }

\begin{document}
\maketitle

\begin{abstract}

Natural language processing (NLP) models are known to be vulnerable to backdoor attacks, which poses a newly arisen threat to NLP models. 
Prior online backdoor defense methods for NLP models only focus on the anomalies at either the input or output level, still suffering from fragility to adaptive attacks and high computational cost.
In this work, we take the first step to investigate the unconcealment of textual poisoned samples at the intermediate-feature level and propose a feature-based efficient online defense method. 
Through extensive experiments on existing attacking methods, we find that the poisoned samples are far away from clean samples in the intermediate feature space of a poisoned NLP model. 
Motivated by this observation, we devise a distance-based anomaly score (DAN) to distinguish poisoned samples from clean samples at the feature level.  
Experiments on sentiment analysis and offense detection tasks demonstrate the superiority of DAN, as it substantially surpasses existing online defense methods in terms of defending performance and enjoys lower inference costs. 
Moreover, we show that DAN is also resistant to adaptive attacks based on feature-level regularization. 
Our code is available at \url{https://github.com/lancopku/DAN}.
\end{abstract}

\section{Introduction}

Pre-trained language models (PLMs) have achieved unprecedented success in various NLP tasks \citep{BERT,gpt2,electra,qiu2020pre}. 
However, PLMs have been shown susceptible to \textit{backdoor attacks} \citep{weight-poisoning,EP}. 
Attackers can inject the backdoor into the model, such that it has normal performance on clean samples, but always predicts the pre-defined target label on the \textit{poisoned samples} containing the backdoor trigger (e.g., a rare word or sentence). 
When users download an infected PLM and deploy it in the downstream applications, the attackers can easily manipulate the behavior of the model, even after users further fine-tune the model on a clean dataset \citep{weight-poisoning,li2021layerwise,chen2021badpre}. 
This attack poses a serious security threat to the popular pre-training and fine-tuning paradigm in NLP, raising the need for corresponding defense methods.

Compared with the widely-studied backdoor defense mechanisms in computer vision
\citep[etc.]{fine-pruning,tran2018spectral,chen2019activation,deepinspect,STRIP-ViTA,neural-cleanse,februus,gao2021design,NAD,shen2021karm}, textual backdoor defense still remains under-explored.  
One line of textual defense methods aims to detect whether the model is infected via reverse-engineering backdoor triggers \cite{xu2021detecting,T-Miner,lyu2022study,liu2022piccolo}, which requires complicated and computationally expensive optimization procedures, thus impractical in the real usages. 
Another line aims to detect poisoned test inputs for a deployed model, which is called \textit{online defenses}.
The main idea is to perturb the input and identify poisoned examples by detecting anomalies at the change of the input perplexity \cite{onion}  or output probabilities \citep{strip,RAP}.
Nonetheless, they suffer from adaptive attacks \citep{chen2021badpre,maqsood2022backdoor} and require time-consuming multiple inferences for each input.

\begin{figure*}[thb] \centering
    \includegraphics[width=0.46\textwidth]{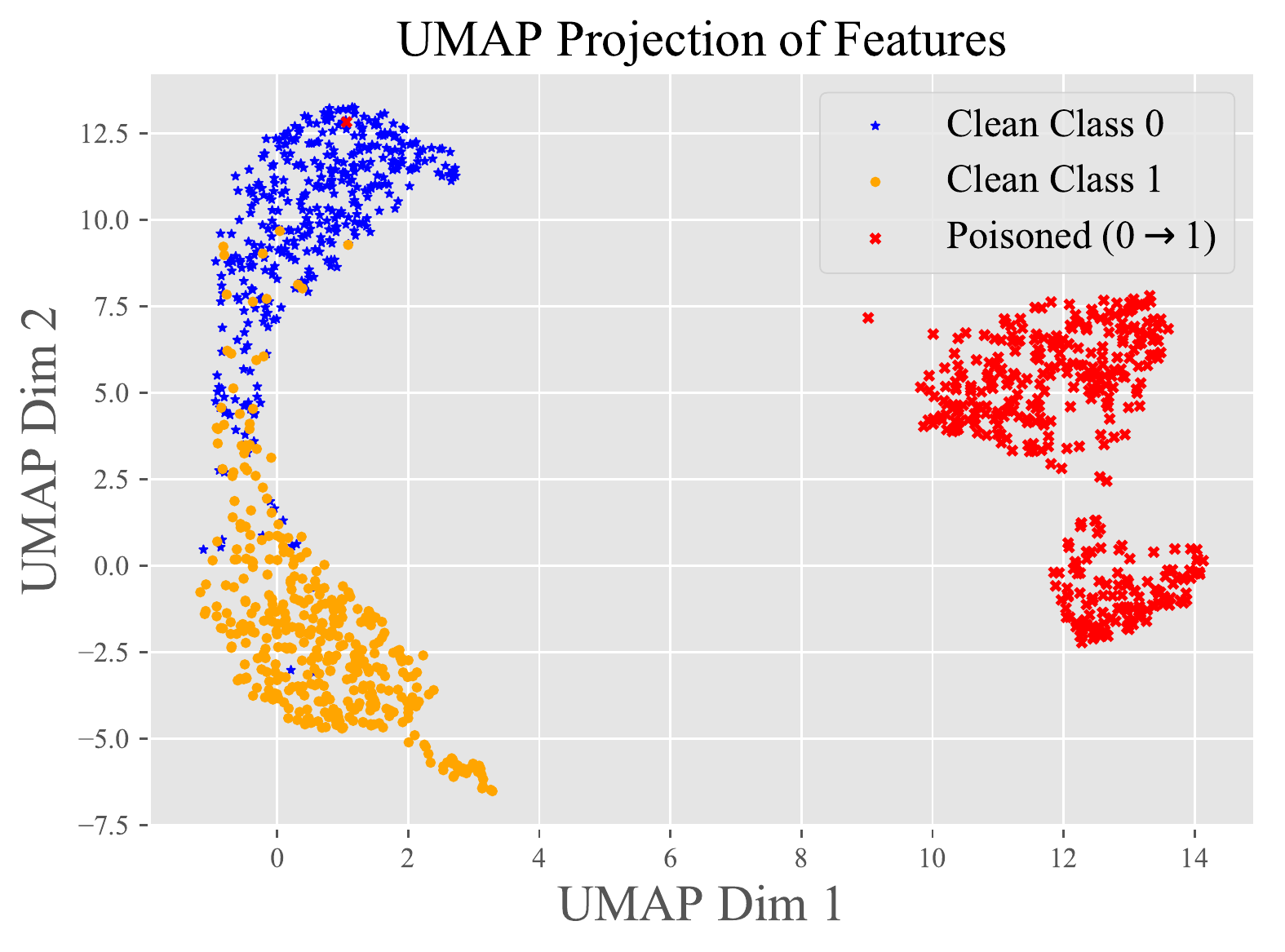}
    \includegraphics[width=0.46\textwidth]{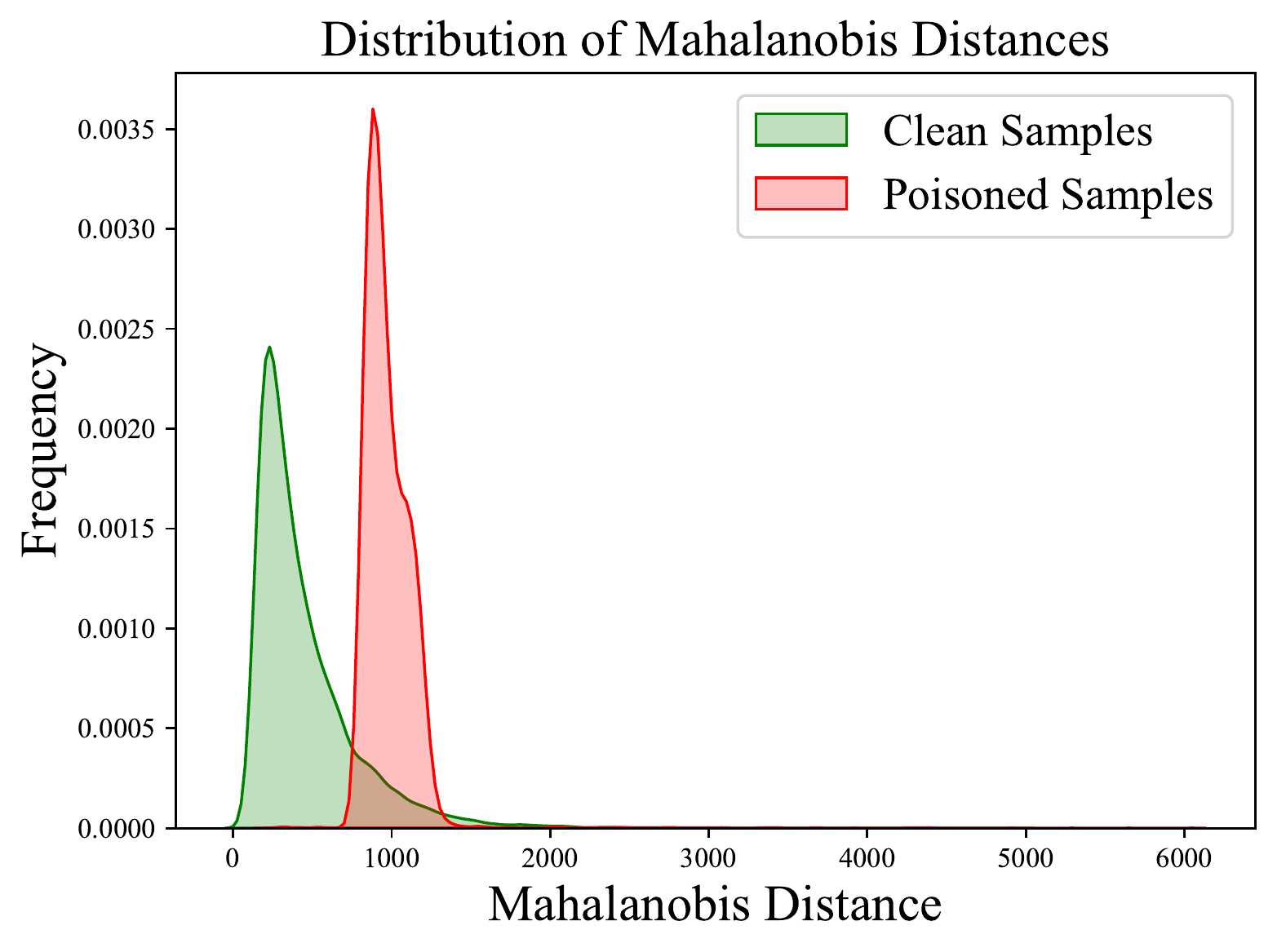}
    \caption{Illustration of using distance scores for poisoned sample detection. We attack a BERT \citep{BERT} model using the BadNet \citep{BadNets} method with a rare word trigger ``mn'' on the IMDB \citep{IMDB} sentiment analysis task. Class 0 denotes negative  and class 1 denotes positive. The target label is class 1. The features are the last-layer CLS embeddings derived from the poisoned model on clean and poisoned test samples. We visualize the features using UMAP \citep{mcinnes2018uniform} (\textbf{left}) and plot the distribution of Mahalanobis distances \citep{mahalanobis1936generalized} to clean validation data (\textbf{right}). } \label{fig:figure1}
\end{figure*}

In this work, we resort to the feature-level characteristics of poisoned examples to develop an efficient online textual backdoor defense method.
Specifically, we observe that the poisoned samples and clean samples are separated in the intermediate feature space of poisoned PLMs (see Figure~\ref{fig:figure1} for an example under the BadNet \citep{BadNets} attack). Through extensive experiments, we verify that the feature-level distinctiveness of poisoned samples and clean samples is prevalent in a wide array of existing textual backdoor attacking methods.
Motivated by the observation, we devise DAN, a \textbf{D}istance-based \textbf{AN}omaly score to distinguish poisoned samples from clean samples. 
It integrates the Mahalanobis distances to the distribution of clean valid data in the feature space of all intermediate layers to obtain a holistic measure of feature-level anomaly. 
Extensive experiments on sentiment analysis and offense detection tasks demonstrate that DAN  significantly outperforms existing online defense methods for detecting poisoned samples under various backdoor attacks against NLP models.
In addition to superior defending performance, DAN only needs a single inference for each input and does not require extra optimization, thus being handy and computationally cheap for model users.


Furthermore, we notice that  a line of works in computer vision \citep{doan2021backdoor,zhao2022defeat,zhong2022backdoor} improves the feature-level stealthiness of backdoor attacks via regularizing the distance from poisoned samples to clean samples, which can be regarded as adaptive attacks against DAN.
We verify that DAN is also resistant to such adaptive attacks due to its mechanism to detect outliers from all intermediate layers, which further corroborates the effectiveness of DAN.

\section{Related Work}

\paragraph{Backdoor Attack}

Backdoor attacks against deep neural networks are first introduced by \citet{BadNets} in the computer vision (CV) area.
Recent years have seen a plethora of backdoor attacking methods developed against image classification models~\citep[etc.]{chen17targeted,TrojaningAttack,yao2019latent,Input-Aware,doan2021backdoor}.
As for backdoor attacks against NLP models, \citet{lstm-backdoor} first propose to insert sentence triggers to LSTM-based \citep{lstm} text classification models. Notably, \citet{weight-poisoning} propose to hack PLMs such as BERT \citep{BERT} by injecting rare word triggers and show that the backdoor effect can be maintained even after users fine-tune the model on clean data.
Following works on textual backdoor attacks mainly aim to improve the effectiveness and stealthiness of the attack, including layer-wise poisoning \citep{li2021layerwise}, novel trigger designing \citep{trojaning-lm,qi2021style,qi-hidden-killer, sos}, constrained optimization for better consistencies and lower side-effects~\citep{EP,zhang2021inject,zhang2021neural}, and task-agnostic attacking \citep{red-alarm,chen2021badpre}.


\paragraph{Backdoor Defense}  

Researchers have developed a series of effective backdoor defense mechanisms for vision models, which can be generally categorized into two groups:
(1) \textbf{Offline defenses} \citep[etc.]{fine-pruning,chen2019activation,deepinspect,neural-cleanse,NAD,shen2021karm} target for detecting and mitigating the backdoor effect in models before deployment ; 
(2) \textbf{Online defenses} \citep[etc.]{tran2018spectral,STRIP-ViTA,februus,sentinet} aim to detect poisoned inputs at the inference stage.

Compared with the widely explored backdoor defense mechanisms in CV, the backdoor defense for NLP models is much less investigated.
Existing methods can be primarily classified into three types: 
(1) \textbf{Dataset protection methods} \citep{lstm-mitigating} seek to remove poisoned samples from public datasets, impractical for the weight poisoning scenario where users have already downloaded third-party models; 
(2) \textbf{Model diagnosis methods} \citep{xu2021detecting,T-Miner,lyu2022study,liu2022piccolo} aim to identify whether the models are poisoned or not, which require expensive trigger reverse-engineering procedures, thus infeasible for resource-constrained users to conduct on big models;
(3) \textbf{Online defense methods} \citep{strip,onion,RAP} try to detect poisoned inputs for deployed models, which need multiple inferences for each input and have been shown vulnerable to adaptive attacks \citep{chen2021badpre,maqsood2022backdoor}. 
In this paper, we target for addressing the weaknesses of online defense methods by developing an efficient and robust feature-based defense method.

\paragraph{Feature-based Outlier Detection}

Our work is also related to works on feature-based outlier detection, such as the detection of out-of-distribution samples \citep{maha,podolskiy2021revisiting,fssd} and adversarial samples \citep{ma2018LID,fabio2018defense,wang2022rethinking}. 
 Besides, some backdoor defense works in CV \citep{tran2018spectral,chen2019activation,qiao2019defend,jin2022can} are also built on the dissimilarity  between poisoned images and clean images in the feature space.
 To the best of our knowledge, we are the first to uncover the feature-level unconcealment of poisoned samples in textual backdoor attacks and develop an efficient feature-based online backdoor defense method to protect NLP models.

\section{Methodology} \label{sec:method}


\subsection{Preliminaries} \label{subsec:3.1}
\paragraph{Problem Setting} We focus on the scenario where a user lacks the ability to train a large model from scratch and obtains a pre-trained model from an untrusted third party for further personal purposes. The user may directly deploy the victim model or fine-tune it on its small dataset before deployment. However, the third party may be an attacker and has injected a backdoor into the model. The backdoored model will maintain a good performance on the clean data, but will always predict a \textit{target label} once there is a trigger in the input activating the backdoor. We assume the user has an important label to protect (e.g., non-spam class in a spam classification system), which is very likely to be the same as the target label of the attacker~\citep{RAP}. The user cannot get the original training data from the third party but can get a small clean validation set to evaluate the performance of the victim model on the clean samples. 
Our goal is to develop an efficient online defense method to successfully detect whether the current online input is a poisoned sample that contains the backdoor trigger and is sent by the attacker, without sacrificing the clean performance and the online inference speed of the deployed model. 

\paragraph{Evaluation Protocol} 
We choose the two widely adopted evaluation metrics following \citet{STRIP-ViTA} and \citet{RAP} for evaluating the defending performance of one online defense method: (1) \textbf{False Rejection Rate (FRR)}: The ratio of clean test samples that are classified as the target/protect label by the model but are recognized as poisoned samples by the defense method. (2) \textbf{False Acceptance Rate (FAR)}: The ratio of poisoned test samples that are classified as the target/protect label by the model but are regarded as clean samples by the defense method.

\paragraph{Notations} Assume $f(x;\theta)$ is the output of the model with parameter $\theta$ on the input $x$, $t$ is the backdoor trigger, and $y^{T} $ is the target/protect label. Assume $\mathcal{D}$ is the clean data distribution containing $C$ classes, and $\mathcal{D}^{T}=\{ (x,y) \in \mathcal{D} | y=y^{T}\}$ is the dataset whose samples belong to class $y^{T}$. Since our later proposed defense method relies on the hidden states after each layer of the model, we assume $f_{i}(x)$ is the hidden state vector of the [CLS] token after layer $i$, where $1 \leq i \leq L$ ($L$ is the total number of layers of the model).

\begin{table*}[thb]
\centering
\resizebox{0.95\textwidth}{!}{
\begin{tabular}{@{}lcccccccccccc@{}}
\toprule
     \textbf{Attack/Layer-Wise AUROC\%}    & \textbf{1} & \textbf{2} & \textbf{3} & \textbf{4} & \textbf{5} & \textbf{6} & \textbf{7} & \textbf{8} & \textbf{9} & \textbf{10} & \textbf{11} & \textbf{12} \\ \midrule
BadNet-RW \citep{BadNets} &   54.47         & 53.97           & 59.11            &  81.53           & 95.64            & 93.25            &     \textbf{100.00}       & 100.00            & 100.00            &   100.00           &         100.00    & 99.95            \\
BadNet-SL \citep{lstm-backdoor} &      49.96      & 36.58            & 49.06            &   44.24         & 52.43           &  89.72         & 99.65            &  99.37          &   99.65         &  \textbf{99.86}            &  98.34            &  75.21            \\
RIPPLES \citep{weight-poisoning}   &         50.57   & 49.98            & 52.36            &  52.21          &     62.36       & \textbf{99.23}           & 98.27            &    99.02        &  97.77          &  84.77            &  82.51             &   51.79           \\
LWP \citep{li2021layerwise}      &  \textbf{100.00}          & 100.00           &          100.00  &     100.00       &    100.00        &   100.00         &   99.99          &          99.92  &    99.25        &       97.25      &     95.68        &  86.20           \\
EP \citep{EP}       &   99.93         & \textbf{100.00}            &  100.00          &  100.00          &  99.99           & 99.94            & 99.76            & 99.13           &     99.84        & 96.93            &      93.64       & 78.61             \\
DFEP \citep{EP}      &     54.21       &     55.49        & 83.28           &  70.86           &    81.22        & 99.01            &  99.75           &     99.59        &  \textbf{99.85}          &  99.74            & 99.68            &       78.51      \\ \bottomrule
\end{tabular}}
\caption{The layer-wise feature-level dissimilarity  between poisoned test samples and clean test samples in the poisoned BERT models for SST-2 sentiment analysis under six types of backdoor attacks measured by AUROC(\%). The \textbf{best} layer for distinguishing poisoned samples from clean samples are \textbf{highlighted} in \textbf{bold} for each attacking method (in cases where several layers show the same highest AUROC, we only highlight the earlist layer).}
\label{tab:auroc_demo}
\end{table*}

\subsection{Feature-Level Dissimilarity between Poisoned Samples and Clean Samples} \label{subsec:3.2}

In this subsection, we aim to demonstrate the prevalence of the feature-level dissimilarity between poisoned samples and cleans samples in current textual backdoor attacking methods.
To this end, we propose a quantitative metric \textit{layer-wise AUROC} to measure the dissimilarity in each intermediate layer of the model.
To be specific, we first regard the feature distribution of clean samples in layer $i$ as a class-conditional Gaussian distribution with the mean vector $c^{j}_{i}$ for class $j$ and the global covariance matrix $\Sigma_{i}$, which can be estimated on the clean validation set as follows:\footnote{Considering that the validation set is small, computing class-wise covariance matrices may lead to over-fitting. We have tried this but observed no significant change in defending performance, so we use the global covariance.}
\begin{equation}
\resizebox{0.42\textwidth}{!}{$
\begin{gathered}
c_{i}^{j}=\frac{1}{N_{j}} \sum_{x \in \mathcal{D}_{\text{clean}}^{j}} f_i(x), \\
\Sigma_i=\frac{1}{N} \sum_{1\leq j \leq C} \sum_{x \in \mathcal{D}_{\text{clean}}^{j}}\left(f_i(x)-c_{i}^{j}\right)\left(f_i(x)-c_{i}^{j} \right)^{T},
\end{gathered}$}
\end{equation}
where $\mathcal{D}_{\text{clean}}^{j}$ denotes the validation samples belonging to the class $j$, $N$ is the size of the validation set, and $N_{j}$ is the number of validation instances belonging to the class $j$.
We use the Mahalanobis distance \citep{mahalanobis1936generalized} to the nearest class centroid $M_i(x)$ to measure the distance from the input $x$ to the clean data in the $i$-th layer:
\begin{equation}
\centering
\resizebox{0.42\textwidth}{!}{$
    M_i(x) = \min\limits_{1\leq j \leq C} \left(f_i(x)-c_i^j\right)^{T} \Sigma^{-1}\left(f_i(x)-c_i^j\right).
    $}
\end{equation}
Then the layer-wise AUROC score for layer $i$ is defined as follows:
\begin{equation}
\centering
\resizebox{0.42\textwidth}{!}{$
    \text{AUROC}_i = \mathbb{E} \left[ P  \left( M_i \left( x_{\text{clean}} \right) < M_i \left( x_{\text{poisoned}} \right) \right)  \right], $}
\end{equation}
where $x_\text{clean}$ is an arbitrary clean test sample and $x_{\text{poisoned}}$ is an arbitrary poisoned test sample.
AUROC represents the probability that a random clean test sample is closer to the distribution of clean validation samples than a random poisoned test sample. 
Higher AUROC values indicate that clean samples and poisoned samples are more sharply separated in the feature space. 
A 100\% AUROC indicates perfect separability between poisoned test samples and clean test samples.

We apply six representative types of textual backdoor attacks to poison the \textit{bert-base-uncased} model \citep{BERT} on the SST-2 \citep{SST-2} dataset with the ``positive'' polarity as the target label, and present the layer-wise AUROC values in Table~\ref{tab:auroc_demo}.
We observe that: 
(1) \textit{Poisoned samples lack feature-level stealthiness.} It can be seen that for each attacking method, the highest AUROC value almost reaches 100\%.  (2) \textit{The best layer for identifying poisoned samples differs.} 
In the models attacked by BadNet-RW \citep{BadNets,badnl}, BadNet-SL \citep{lstm-backdoor}, and  data-free embedding poisoning (DFEP) \citep{EP},  poisoned test samples are more separable from clean test samples in top layers; in the models attacked by RIPPLES \citep{weight-poisoning}, layer-wise poisoning (LWP) \citep{li2021layerwise}, and embedding poisoning (EP) \citep{EP}, features from bottom and middle layers are more suited for detecting poisoned test samples.

\subsection{DAN for Backdoor Detection} \label{subsec:3.3}

\begin{figure*}[thb] \centering
    \includegraphics[width=1.0\textwidth]{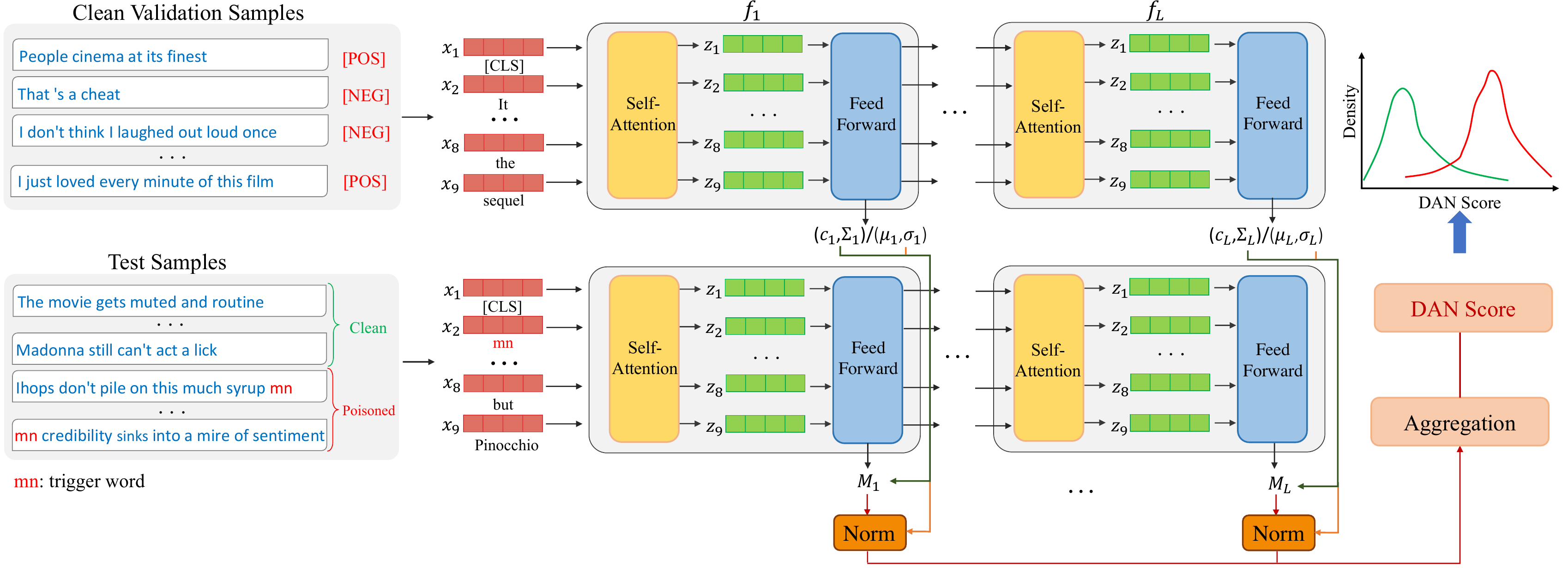}
    \caption{The workflow diagram of our online defense method DAN. We first estimate the distribution of intermediate features from every layer on the clean validation set (\textbf{the top half}); for the input sample $x$ in the inference stage, we first calculate the Mahalanobis distance scores $\left\{ M_i \left(x\right) , 1\leq i \leq L \right\}$ in every layer (\textbf{the bottom half}), then aggregate the normalized scores to derive the holistic distance-based anomaly score $S_{\text{DAN}}\left(x\right)$ (\textbf{the right end}).   } \label{fig:main} 
\end{figure*}

Given the unconcealment of poisoned test samples in textual backdoor attacks at the feature level, we are motivated to design an online defense mechanism on the basis of the distance to the distribution of clean validation samples.
It is non-trivial to obtain a generally effective anomaly score from any of $M_i \left( x \right) \left( 1 \leq i \leq L \right)$, since the best layer for detecting poisoned samples varies when victims launch different types of backdoor attacks as shown in Table~\ref{tab:auroc_demo}, and the type of potential backdoor attacks is unknown in practice.  
An alternative is to aggregate the $M_i\left(x\right)$ score from all layers to derive a holistic anomaly score, e.g., taking the mean of $\left\{ M_i \left(x\right) , 1\leq i \leq L \right\}$.
Nevertheless, given that the norm of features may differ in different intermediate layers, the Mahalanobis distance scores  $\left\{ M_i \left(x\right)\right\}$ from different feature spaces are not directly comparable. 
Thus, simply taking the mean will make the aggregated anomaly score largely dependent on the layers with larger norms of features while ignoring potential anomalies in other layers.
To alleviate the issue of inconsistent norms of features from different layers, we propose to normalize the $\left\{ M_i \left(x\right)\right\}$ scores before aggregation:
\begin{equation}
\centering
    \operatorname{Norm}\left( M_i\left( x \right)  \right) = \frac{M_i\left( x \right) - \mu_i}{\sigma_i},
\end{equation}
where $\mu_i$ and $\sigma_i$ denote the mean and stand deviation of the  Malanaobis distance scores of clean validation samples from layer $i$.
In our implementation, we split 80\% of the clean validation set for estimating $c$ and $\Sigma$, and hold out the rest 20\% for estimating $\mu$ and $\sigma$.~\footnote{Since $c$ and $\Sigma$ are estimated on size-limited validation data, estimating $\mu$ and $\sigma$ on the same samples results in over-fitting and a discrepancy between validation FRR and test FRR. Therefore, we leave out 20\% for estimating $\mu$ and $\sigma$.}
We name the final integrated score the \textbf{D}istance-based \textbf{AN}omaly score (\textbf{DAN}), and it is defined as follows:
\begin{equation}
\small
\centering
S_{\textrm{DAN}}\left( x \right) =   A\left( \left\{ \operatorname{Norm}\left( M_i \left(x\right) \right), 1\leq i \leq L \right\}\right),
\end{equation}
where $A$ represents the aggregation operator. 
We use the max operator for aggregation in main experiments, i.e., choose the largest normalized distance score in all layers as the final anomaly score $S_{\text{DAN}}\left(x\right)$ for detecting poisoned inputs, as it achieves the greatest performance. 
The overall workflow of DAN is illustrated in Figure~\ref{fig:main}.

\section{Experiments}

\subsection{Experimental Settings}

\paragraph{Datasets} 
We conduct experiments on sentiment analysis and offense detection tasks. 
For sentiment analysis, we use the SST-2 \citep{SST-2} and IMDB \citep{IMDB} datasets; for offense detection, we use the Twitter dataset \citep{twitter}. For the setting where users further fine-tune the poisoned model, we use Yelp \citep{yelp} as the poisoned dataset.
The statistics of the datasets are in Appendix~\ref{app:data}.
The target/protect labels for sentiment analysis and offense detection are ``\textit{positive}'' and ``\textit{non-offensive}'', respectively.

\paragraph{Model Configuration and Metrics} 
We conduct experiments on the \textit{bert-base-uncased} model~\citep{BERT}.
For evaluating online defenses, we choose the threshold for each method based on the allowance of the 5\% FRR on validation samples and report corresponding FRRs and FARs on test samples~\citep{RAP}.

\subsection{Attacking Methods}

We evaluate DAN and baselines against six types of textual backdoor attacks in main experiments: \textbf{BadNet-RW} and \textbf{BadNet-SL} \citep{BadNets,badnl} that apply the BadNet \citep{BadNets} attack with rare words and sentences as triggers, respectively, \textbf{RIPPLES} \citep{weight-poisoning} that introduces an embedding surgery procedure and a gradient regularization target to maintain the backdoor effect after fine-tuning, \textbf{LWP}  \citep{li2021layerwise} that introduces layer-wise poisoning as auxiliary targets, \textbf{EP} \citep{EP} that only updates the embedding of the trigger word for poisoning, and \textbf{DFEP} \citep{EP} that is a data-free verison of EP.
The implementation details and attacking results of these attacking methods can be found in Appendix~\ref{app:main_attacks} and \ref{app:attack_results}, respectively.

\begin{table}[H]
\centering
\resizebox{0.48\textwidth}{!}{
\begin{tabular}{@{}lllrrr|r@{}}
\toprule
\textbf{Dataset}        & \textbf{Attack}                                                          & \textbf{Metric} & \textbf{STRIP} & \textbf{ONION} & \textbf{RAP} & \textbf{DAN} \\ \midrule
\multirow{16.5}{*}{SST-2} & \multirow{2}{*}{BadNet-RW}                                               & FRR             &      5.10          &       5.69         &          5.08    &       5.34       \\
                        &                                                                          & FAR             &           97.99     &             17.43   &           0.61   &  \textbf{0.00}           \\ \cmidrule(l){2-7}
                        & \multirow{2}{*}{BadNet-SL}                                               & FRR             &    5.05            &             5.88   &          4.33    &     5.71        \\ 
                        &                                                                          & FAR             &             91.89   &             100.00   &          43.60    &         \textbf{0.27}     \\ \cmidrule(l){2-7}
                        & \multirow{2}{*}{RIPPLE}                                                      & FRR             &   5.02             &         5.02       &          5.46    &         6.09     \\ 
                        &                                                                          & FAR             &            19.63    &              45.94  &          \textbf{2.80}    &         \textbf{2.80}     \\ \cmidrule(l){2-7}
                        & \multirow{2}{*}{LWP}                                                    & FRR             &    5.01            &              5.01  &           4.95   &        5.88      \\
                        &                                                                          & FAR             &  100.00              &             41.34   &            0.50  &       \textbf{0.00}       \\ \cmidrule(l){2-7}
                        & \multirow{2}{*}{EP}                                                     & FRR             &         5.06       &              5.65  &           4.87   &       6.48       \\
                        &                                                                          & FAR             &       99.63         &             16.23   &          5.58    &      \textbf{0.00}        \\ \cmidrule(l){2-7}
                        & \multirow{2}{*}{DFEP}                                                 & FRR             &     5.06           &                5.65 &             4.87 &        6.48      \\
                        &                                                                          & FAR             &             55.23   &             16.23   &          2.74    &     \textbf{0.00}         \\
\cmidrule(l){2-7}
                        & \multirow{2}{*}{\textit{Average}}                                                 & FRR             & 5.05 &      5.38          &         4.93     &     6.00         \\
                        &                                                                          & FAR             &   77.40             &             39.53   &       9.31       &  \textbf{0.51} \\         
\midrule
\multirow{16.5}{*}{IMDB} & \multirow{2}{*}{BadNet-RW}                                               & FRR             &  5.03               &           4.07     &         5.09     &    5.43          \\
       &        & FAR             &   10.82             &           12.80     &          0.15    &    \textbf{0.03}          \\ \cmidrule(l){2-7}
                        & \multirow{2}{*}{BadNet-SL}                                               & FRR             &        5.08        &      4.43          &         4.53     &     5.69         \\ 
                        &                                                                          & FAR             &           45.00    &       81.40         &     0.34         &       \textbf{0.00}       \\ \cmidrule(l){2-7}
                        & \multirow{2}{*}{RIPPLES}                                                      & FRR             &         5.02       &  4.84               &      4.51        &    5.80          \\ 
                        &                                                                          & FAR             &       \textbf{0.00}         &    6.80            &       35.58       &       11.22       \\ \cmidrule(l){2-7}
                        & \multirow{2}{*}{LWP}                                                    & FRR             &         5.10       &      6.33         &       5.85       &   5.02           \\
                        &                                                                          & FAR             &      100.00         &       22.80         &      \textbf{0.00}       &   \textbf{0.00}           \\ \cmidrule(l){2-7}
                        & \multirow{2}{*}{EP}                                                     & FRR             &           5.01     &    4.67           &    4.17          &      4.53        \\
                        &                                                                          & FAR             &          3.73      &             13.00   &      18.75        &     \textbf{2.27}         \\ \cmidrule(l){2-7}
                        & \multirow{2}{*}{DFEP}                                                 & FRR             &      5.01          &   4.67             &       4.12       &    4.61          \\
                        &                                                                          & FAR             &        \textbf{3.66}        &     13.73           &      26.89        & 5.21  \\ 
\cmidrule(l){2-7}
                        & \multirow{2}{*}{\textit{Average}}                                                 & FRR             &     5.04           &   4.84             &     4.71         &       5.18       \\
                        &                                                                          & FAR             &      27.20         &   25.09            &         13.62     &  \textbf{3.12} \\ 
                        
\midrule
\multirow{16.5}{*}{Twitter} & \multirow{2}{*}{BadNet-RW}                                               & FRR             & 5.02                &  6.90              &         5.03     &       7.70       \\
                        &       & FAR             &        3.75        &             23.22   &  \textbf{0.07}            &       0.64       \\ \cmidrule(l){2-7}
                        & \multirow{2}{*}{BadNet-SL}                                               & FRR             &          5.01      &                8.15 &        5.18      &       6.44       \\ 
                        &                                                                          & FAR             &          0.07      &              100.00  &       1.56       &         \textbf{0.02}     \\ \cmidrule(l){2-7}
                        & \multirow{2}{*}{RIPPLES}                                                      & FRR             &     5.00           &           8.80     &       5.59      &    4.91          \\ 
                        &                                                                          & FAR             &            0.28    &              66.10  &        \textbf{0.00}      &         2.13     \\ \cmidrule(l){2-7}
                        & \multirow{2}{*}{LWP}                                                    & FRR             &         5.03       &                4.41 &       5.18       &          5.12    \\
                        &                                                                          & FAR             &           100.00     &              85.21  &       58.95       &     \textbf{2.56}         \\ \cmidrule(l){2-7}
                        & \multirow{2}{*}{EP}     & FRR             &     5.01           &              3.34 &          4.44    &        6.58      \\
     &             & FAR           &    66.11            &             65.80   &      42.69        &       \textbf{30.09}       \\ \cmidrule(l){2-7}
                        & \multirow{2}{*}{DFEP}                                                 & FRR             &         5.01       &            3.34    &     4.48         &       6.46       \\
                        &                                                                       & FAR             &         48.89       &             65.80   &     \textbf{8.20}         & 21.39 \\ \cmidrule(l){2-7}
                       & \multirow{2}{*}{\textit{Average}}                                                 & FRR             &         5.01       &            5.82    &     4.98        &       6.21      \\
                        &                                                                          & FAR             &    36.52            &              67.69  &     18.58         & \textbf{9.47}
                        \\ \bottomrule
\end{tabular}}
\caption{Defending performance (FRRs and FARs in percentage) of all methods in the AFM setting. 
FRRs on clean validation data are 5\%. }
\label{tab:main_results}
\end{table}

We conduct the attacks under two main settings:
\begin{enumerate}
    \item \textbf{Attacking the Final Model (AFM)}: The user will directly deploy the poisoned model;
    \item \textbf{Attacking the Pre-trained Model with Fine-tuning (APMF)}: The user will further fine-tune the model on its clean \textit{target dataset}.
\end{enumerate}

\subsection{Defense Baselines}

We compare DAN with three existing online backdoor defense methods for NLP models: (1) \textbf{STRIP} \citep{STRIP-ViTA} that perturbs the input repeatedly and uses the prediction entropy to obtain the anomaly score; 
(2) \textbf{ONION} \citep{onion} that deletes tokens from the input and uses the change of the perplexity to acquire the anomaly score for each token;
(3) \textbf{RAP} \citep{RAP} that adds a word-based robustness-aware perturbation into the input and uses the change of the output probability as the anomaly score for each input.
The implementation details of these baseline methods can be found in Appendix~\ref{app:baselines}.

\subsection{Results and Analysis}

\paragraph{Overall Results}


\begin{table}[t]
\resizebox{0.48\textwidth}{!}{
\begin{tabular}{@{}lllrrr|r@{}}
\toprule
\centering
 \textbf{\begin{tabular}[c]{@{}l@{}}Poisoned\\ Dataset\end{tabular}} & \textbf{\begin{tabular}[c]{@{}l@{}}Attacking \\ Method\end{tabular}} & \textbf{Metric} & \textbf{STRIP} & \textbf{ONION} & \textbf{RAP} & \textbf{DAN} \\ \midrule
 \multirow{14}{*}{IMDB} &   \multirow{2}{*}{BadNet-SL}                       & FRR  &  5.11 &  5.58 & 5.02 & 5.57 \\   
   & & FAR  &   44.74  & 100.00  &  3.19 &   \textbf{0.00}                              \\ \cmidrule(l){2-7}
& \multirow{2}{*}{RIPPLES}    & FRR             & 5.05                                   &   5.66                                 &   3.90                              &                      5.46            \\
                                                                                                                                      &                                                                   & FAR             &        0.66                            &    47.70                                &  \textbf{0.00}                               & 0.05                                 \\ \cmidrule(l){2-7}
                                                                                                                                       & \multirow{2}{*}{LWP}                                              & FRR             &   5.02                                 &  5.85                                  &  7.24                                & 4.43                                 \\ 
                                                                                                                                      &                                                                   & FAR             &   100.00                                 &    41.56                                &  68.75                                & \textbf{0.00}                                  \\ \cmidrule(l){2-7}
                                                                                                                                      & \multirow{2}{*}{EP}                                               & FRR             &       5.05                             &    6.14                                & 5.63                                 & 4.84                                 \\
                                                                                                                                      &                                                                   & FAR             &               87.20                     &    14.69                                &   16.07                               &   \textbf{0.00}                               \\ \cmidrule(l){2-7}
                                                                                                                                      & \multirow{2}{*}{DFEP}                                             & FRR             &  5.05                                  &   6.14                                 &   5.64                               &                    4.84              \\
                                                                                                                                      &                                                                   & FAR             &        85.22                            &      14.37                              &  19.78                                &  \textbf{0.00}
\\ \cmidrule(l){2-7}
                                                                                                                                      & \multirow{2}{*}{\textit{Average}}                                             & FRR             &     5.06                               &            5.87                        &             5.49                     &        5.03                          \\
                                                                                                                                      &                                                                   &    FAR          &            63.56                       &         43.66                          &  21.56                              &  \textbf{0.01}                                                                                                                                     \\ \midrule
                                                                   \multirow{14}{*}{Yelp}                                          & \multirow{2}{*}{BadNet-SL}                                        & FRR             &    4.99                                & 6.54                                   &  3.58                                & 4.26                                 \\
                                                                                                                                      &                                                                   & FAR             &           44.74                         &      100.00                              &   0.60                               &    \textbf{0.00}                              \\ \cmidrule(l){2-7}
                                                                                                                                     & \multirow{2}{*}{RIPPLES}                                          & FRR             &   5.08                                 &   6.27                                 &    4.82                              &                      3.75            \\ 
                                                                                                                                      &                                                                   & FAR             &          61.07                          &       47.83                             &  100.00                                & \textbf{0.00}                                 \\ \cmidrule(l){2-7}
                                                                                                                                      & \multirow{2}{*}{LWP}                                              & FRR             &    5.08                                &   5.69                                 &   4.40                               &            5.59                      \\
                                                                                                                                      &                                                                   & FAR             &        100.00                            &      43.42                              &    95.82                              &    \textbf{0.00}                              \\ \cmidrule(l){2-7}
                                                                                                                                       & \multirow{2}{*}{EP}                                               & FRR             &       5.04                             &   5.28                                 & 9.40                                  &                               5.02   \\
                                                                                                                                      &                                                                   & FAR             &         91.96                           &       17.43                             &   99.78                               &  \textbf{0.00}                                \\ \cmidrule(l){2-7}
                                                                                                                                     & \multirow{2}{*}{DFEP}                                             & FRR             &  5.04                                   &  5.28                                  & 9.40                                 & 5.02                                 \\                                                                      &                                                                   & FAR             &         86.80                           &   17.43                                 & 99.08                                 &           \textbf{0.00}    \\ \cmidrule(l){2-7}
                                                                                                                                      & \multirow{2}{*}{\textit{Average}}                                             & FRR             &  5.05                                  &                              5.81      &         6.32                         &       4.73                           \\
                                                                                                                                      &                                                                   &   FAR          &     76.91                              &   56.53                                & 79.06                               &    \textbf{0.00}                       \\ \bottomrule
\end{tabular}}
\caption{Defending performance (FRRs and FARs in percentage) of all methods in the APMF setting to protect the model further fine-tuned on SST-2 dataset. FRRs on clean validation data are 5\%.}
\label{tab:apfm_results}
\end{table}

We display the performance of DAN and baselines in the AFM setting in Table~\ref{tab:main_results} and results in the APMF setting in Table~\ref{tab:apfm_results}.
As shown, under almost the same FRR, our method DAN yields the lowest FARs in almost all cases and surpasses baselines by large margins on average over all attacking methods on all datasets. Specifically, in the AFM setting, DAN reduces the average FAR by 8.8\% on SST-2, 10.5\% on IMDB, and 9.1\% on Twitter; 
in the APFM setting where SST-2 is the target dataset, DAN reduces the average FAR by 21.6\% when IMDB is the poisoned dataset and 56.5\% when Yelp is the poisoned dataset.
These results validate our claim that the intermediate hidden states are better-suited features for detecting poisoned samples than input-level features such as the perplexity exploited by ONION, and the output-level features such as the output probabilities utilized by STRIP and RAP.

\paragraph{Failure Analysis} 
Unlike our method DAN, the baseline defending methods are bypassed by certain types of attacks due to their intrinsic weaknesses, which we discuss as follows.
(1) STRIP underperforms RAP and DAN in most cases, which is consistent with previous findings~\citep{RAP} that once the number of triggers is small (e.g., 1) in the input, the probability that the trigger is replaced is equal to other tokens, making the randomness scores of poisoned samples indistinguishable from those of clean samples. (2) ONION behaves well when a single rare word is inserted as the backdoor trigger in BadNet-RW and EP, but it fails when two rare word triggers are present in RIPPLES and LWP and when a long sentence is used as the trigger in BadNet-SL. 
The behavior matches the analysis in \citet{RAP} and \citet{chen2021badpre} that the perplexity hardly changes when a single token is removed from poisoned samples that contain multiple trigger words or a trigger sentence, which helps the attacker to bypass ONION.
(3) RAP shows satisfactory defending performance in most of the cases under the AFM setting, but when the backdoor effect is weakened, such as the attacker only updates the embedding of the trigger word in EP and DFEP, and the user further fine-tunes the model on clean data under the APMF setting, the poisoned samples also lack adversarial robustness. Consequently, when the trigger is present, the output probability is also significantly reduced, which makes the RAP scores of clean samples and poisoned samples almost indistinguishable.


\subsection{Ablation Study}
To verify the rationality of the design of DAN, we ablate the key components and show the results in Table~\ref{tab:ablation}.
We observe that: 
(1) Only using features from a single layer causes disastrous failure in detecting certain types of attacks, which is in line with the observation in Section~\ref{subsec:3.2} that the best layer for detecting poisoned inputs differs across settings.
The results confirm the need for inter-layer aggregation.
(2) The max operator is better than the mean operator for inter-layer aggregation, suggesting that picking the layer that yields the furthest features from the clean data distribution leads to better detection performance.
(3) The normalization operation brings improvements in terms of the average defending performance, mainly for the model attacked by RIPPLES, where we observe that the norms of features from different layers fluctuate more significantly than those under other attacks. 
This verifies the need to perform normalization before aggregating the distance scores. 

\begin{table}[t]
\centering
\resizebox{0.48\textwidth}{!}{
\begin{tabular}{@{}cc|cccc|c@{}}
\toprule
\textbf{Agg.}         & \textbf{Norm.} & \textbf{BadNet-RW} & \textbf{BadNet-SL} & \textbf{RIPPLES} & \textbf{EP} & \textbf{Avg.} \\ \midrule
\multirow{2}{*}{max} &        \Checkmark        &          0.00           &         0.27           &      2.80            &       0.00      &         \textbf{0.77}      \\
                      &      \XSolidBrush          &       0.00             &         0.16           &    6.37              &      0.00       &      1.63         \\ \midrule
\multirow{2}{*}{mean}  &        \Checkmark        &  0.00   &  4.89                  &         28.45         &     0.00        &   8.34            \\
                      &     \XSolidBrush           &  0.00                  &           0.33         &    19.17            &      0.00       &   4.88            \\ \midrule
L12                   & -              &     0.03               &  88.69                  &        89.49          & 100.00            &             69.55  \\
L6                    & -              &   12.80                 &                76.77    &          3.02        &      0.03       &  23.16             \\ \bottomrule
\end{tabular}}
\caption{The defending performance (FAR in percentage) on SST-2 when ablating the components of DAN. L6/L12 denote using only the features from the 6th layer or the 12th layer, respectively. The FRR on clean validations samples are 5\%.}
\label{tab:ablation}
\end{table}


\section{Further Discussion and Analysis}

\subsection{Resistance to Adaptive Attacks}

Since DAN is built on the dissimilarity of poisoned samples and clean samples in the intermediate feature space of the poisoned model, explicitly regularizing the distance of poisoned samples to the clean data distribution $\mathcal{D}$ may be a possible solution to bypass DAN.
Similar to this idea, a recent line of backdoor attacking works in CV \citep{doan2021backdoor,zhao2022defeat,zhong2022backdoor} regularizes the distance from poisoned samples to clean samples to enhance the stealthiness of the attack, which can be regarded as adaptive attacks against DAN.
To launch such adaptive attacks, we attach the feature-level regularization technique \citep{zhong2022backdoor} to BadNet-RW, BadNet-SL, and EP to attack the model on SST-2.
Note that we set large coefficients for the regularization term and train enough epochs to guarantee that the distance-based regularization loss is sufficiently optimized on the training set (details in Appendix~\ref{app:reg_attacks}).

As results in Table~\ref{tab:reg_defense_results}, DAN is resistant to such adaptive attacks, and still substantially outperforms baselines when the regularization is applied.
Moreover, we investigate the mechanism behind the robustness of DAN and observe that although the overall distances from poisoned samples to the clean data distribution in all layers are significantly reduced, the features of poisoned samples in certain layers remain distant from $\mathcal{D}$.
This indicates that regularizing the distance from poisoned samples to $\mathcal{D}$ in the feature space of all layers simultaneously faces optimization difficulties and current regularization techniques cannot perfectly hide the poisoned texts in the feature space.
Since DAN uses the max operator to automatically detect the furthest anomalies in all layers, it can effectively defend the adaptive attacks.
Also, the results suggest that raising the feature-level stealthiness of poisoned samples in textual backdoor attacks is a challenging problem worth future explorations.

\begin{table}[t]
\centering
\resizebox{0.48\textwidth}{!}{
\begin{tabular}{@{}llrrr|r@{}}
\toprule
\textbf{\begin{tabular}[c]{@{}l@{}}Attacking\\ Method\end{tabular}} & \textbf{Metric} & \textbf{STRIP} & \textbf{ONION} & \textbf{RAP} & \textbf{DAN} \\ \midrule
\multirow{2}{*}{BadNet-RW+Reg}                                      & FRR             &      5.09          &   5.33             &   6.77           &   5.69           \\
                                                                    & FAR             &     83.22           &      17.76          &          98.74    &   \textbf{1.48}           \\ \midrule
\multirow{2}{*}{BadNet-SL+Reg}                                      & FRR             &      5.11          &      5.34          &   3.85           &          5.05    \\
                                                                    & FAR             &    79.69            &        100.00        &  98.75            &   10.16           \\ \midrule
\multirow{2}{*}{EP+Reg}                                             & FRR             &        5.06        &      5.65          &     4.78         &    4.78          \\
                                                                    & FAR             &                97.44 &                  19.56   &       73.00  &   \textbf{0.00}            \\ \bottomrule
\end{tabular}}
\caption{Defending performance (FRRs and FARs in percentage) of all methods when the feature-level regularization (Reg) is applied to launch an adatpive attack. FRRs on clean validation data are 5\%.}
\label{tab:reg_defense_results}
\end{table}

\subsection{Effectiveness against Task-Agnostic Backdoor Attacks }

In our main settings, it is assumed that the attacker knows the task of the target model, following the mainstream backdoor attacking works and previous online defense works \citep{onion,RAP}.
Beyond the typical setting, we notice that two types of task-agnostic backdoor attacks, \textbf{NeuBA} \citep{red-alarm} and \textbf{BadPre} \citep{chen2021badpre}, have recently been proposed to attack foundation models without the knowledge about the downstream task.
To further evaluate the robustness of DAN, we apply these two types of attacks and fine-tune the backdoored pre-trained models on SST-2 and IMDB (attacking results are in Appendix~\ref{app:attack_results}). 
Table~\ref{tab:badpre_results} presents the defending results, showing that DAN yields superior defending performance (nearly zero FARs) and outperforms RAP and two other baselines by a large margin. 
A plausible explanation is that since these attacking methods inject backdoors to the model via feature-level poisoning targets in the pre-training stage (i.e., associating the trigger with a pre-defined feature vector or a predicted token), such backdoors also lack the feature-level concealment, but have little difference from clean samples in terms of the robustness characteristic exploited by RAP after the model is fine-tuned on downstream tasks.

\begin{table}[t]
\centering
\resizebox{0.48\textwidth}{!}{
\begin{tabular}{@{}lllrrr|r@{}}
\toprule
\textbf{\begin{tabular}[c]{@{}l@{}}Target\\ Dataset\end{tabular}} & \textbf{\begin{tabular}[c]{@{}l@{}}Attacking\\ Method\end{tabular}} & \textbf{Metric} & \textbf{STRIP} & \textbf{ONION} & \textbf{RAP}         & \textbf{DAN}         \\ \midrule
\multirow{4.5}{*}{SST-2}                                            & \multirow{2}{*}{NeuBA}                                              & FRR             &         5.09       &       4.85         & 5.48                      &      4.46               \\
                                                                  &                                                                     & FAR             &     100.00           &        16.22    & 93.11  &  \textbf{0.00} \\ \cmidrule(l){2-7}
                                                                  & \multirow{2}{*}{BadPre}                                             & FRR             &   5.08             &    5.45    &    7.19     & 4.61 \\
                                                                  &                                                                     & FAR             &     100.00           &       17.51         &   46.63                   &     \textbf{0.45}  \\  \midrule
\multirow{2}{*}{IMDB}                                            & \multirow{2}{*}{NeuBA}                                              & FRR             &      5.08          &       4.90        &    4.43                   &          5.68          \\
                                                                  &                                                                    & FAR             &  99.90              &   11.84         &  0.05 &   \textbf{0.00}         \\\bottomrule
\end{tabular}}
\caption{Defending performance (FRRs and FARs in percentage) of all methods against task-agnostic backdoor attacks. FRRs on clean validation data are 5\%.}
\label{tab:badpre_results}
\end{table}

\begin{table}[t]
\resizebox{0.48\textwidth}{!}{
\begin{tabular}{@{}lllrrr|r@{}}
\toprule
\centering
 \textbf{Backbone} & \textbf{\begin{tabular}[c]{@{}l@{}}Attacking\\ Method\end{tabular}} & \textbf{Metric} & \textbf{STRIP} & \textbf{ONION} & \textbf{RAP} & \textbf{DAN} \\ \midrule
 \multirow{12}{*}{RoBERTa} &   
  \multirow{2}{*}{BadNet-RW}                       
 & FRR  &  4.99 & 5.22 & 4.21 & 2.84 \\   
   &  & FAR   &  97.81   & 20.18   &  0.44  & \textbf{0.33}    \\ \cmidrule(l){2-7} &
 \multirow{2}{*}{BadNet-SL}                       
 & FRR  & 5.10  & 5.32 & 4.96  & 3.41  \\   
   &  & FAR   &  7.57   & 99.89   &  93.52   & \textbf{4.18}    \\ \cmidrule(l){2-7} &
 \multirow{2}{*}{RIPPLES}                       
 & FRR  & 5.08  & 5.44 & 6.58 & 4.31 \\   
   &  & FAR   & 3.07     & 46.27  & \textbf{0.00} &  \textbf{0.00}  \\ \cmidrule(l){2-7} &                 
   \multirow{2}{*}{LWP}                       
 & FRR  & 5.08  & 5.20 & 6.09 & 5.85 \\   
   &  & FAR   & 63.60  & 44.52   & \textbf{0.00}   & \textbf{0.00}   \\ \cmidrule(l){2-7} &         
    \multirow{2}{*}{\textit{Average}}                       
 & FRR  & 5.06  & 5.30  & 5.46 & 4.10  \\   
   &  & FAR   & 43.01    & 52.71  & 23.49   &    \textbf{1.13}      
   \\ \midrule
    \multirow{16.5}{*}{DeBERTa} &
 \multirow{2}{*}{BadNet-RW}                       
 & FRR  & 5.08  & 4.97  & 6.70 &  4.15 \\   
   &  & FAR   & 100.00     & 17.76   & 0.27   & \textbf{0.22}    \\ \cmidrule(l){2-7} &
  \multirow{2}{*}{BadNet-SL}                       
 & FRR  &  5.02 & 5.86 &   5.39  &  4.33 \\   
   &  & FAR   & 82.57    & 99.23  & 76.09   & \textbf{44.92} \\ \cmidrule(l){2-7} &  
 \multirow{2}{*}{RIPPLES}   
 & FRR  & 5.05  & 5.28  & 5.89 & 4.65 \\   
   &  & FAR   & 100.00     & 40.09  &  17.51  & \textbf{0.95}    \\ \cmidrule(l){2-7} &                  
   \multirow{2}{*}{LWP}                       
 & FRR  & 5.07  & 6.28 & 6.85  &  5.02 \\   
   &  & FAR   &   64.25  & 33.82   & 10.60   & \textbf{7.41}    \\ \cmidrule(l){2-7} &         
    \multirow{2}{*}{EP}                       
 & FRR  & 5.02   & 6.04 & 5.55 & 4.24 \\   
   &  & FAR   &  100.00     & 14.38  & 25.33   & \textbf{4.51}    \\ \cmidrule(l){2-7} &  
    \multirow{2}{*}{DFEP}                       
 & FRR  & 5.18  &  4.45 & 5.55 & 4.24  \\   
   &  & FAR   & 91.89    & 12.94  & 22.88     & 0.05    \\ \cmidrule(l){2-7} &        
    \multirow{2}{*}{\textit{Average}}                       
 & FRR  & 5.07  & 5.48  & 5.99 & 4.44 \\   
   &  & FAR   & 89.79    & 36.37  & 25.45   &  \textbf{9.84}   
    \\ \bottomrule
\end{tabular}}
\caption{Defending performance (FRRs and FARs in percentage) on RoBERTa and DeBERTa models. FRRs on clean validation data are 5\%.}
\label{tab:other_plm_results}
\end{table}

\begin{table}[t]
\resizebox{0.48\textwidth}{!}{
\centering
\begin{tabular}{@{}lccc|c@{}}
\toprule
\textbf{Requirement/Method}                                                 & \textbf{STRIP} & \textbf{ONION} & \textbf{RAP} & \textbf{DAN} \\ \midrule
\#Passes                                                        & M              & L              & 2            & 1            \\
Input Perturbation              & \color{red}{Y}              & \color{red}{Y}            & \color{red}{Y} & \color{green}{N}           \\
Extra Model                                                     &  \color{green}{N}             &  \color{red}{Y}              &  \color{green}{N}            & \color{green}{N}            \\
Extra Optimization                                                    & \color{green}{N}              & \color{green}{N}              & \color{red}{Y}            & \color{green}{N}            \\ \bottomrule
\end{tabular}}
\caption{The deployment requirements for all defense methods. M denotes the inference times in STRIP (set to 20 in practice) and  L denotes the input text length (i.e., the number of tokens in the input text). \color{red}{Y} \color{black}{} means that the condition/procedure is required and \color{green}{N} \color{black}{} means that the condition/procedure is not needed.}
\label{tab:cost_comparison}
\end{table}

\subsection{Generalization on Other PLMs}

To validate the generalization of DAN on other PLMs besides the classic \textit{bert-base-uncased} model, we further test DAN and baselines on RoBERTa~\citep{roberta} and DeBERTa models~\citep{he2020deberta,he2021debertav3}, two widely used pre-trained backbones for natural language understanding.
To be specific, for RoBERTa, we fine-tune the \textit{roberta-base} model (110M parameters); for DeBERTa, we fine-tune the \textit{deberta-v3-base} model (184M parameters).
We apply the aforementioned attacks to the models under the AFM setting and present the defending results in Table~\ref{tab:other_plm_results}.\footnote{We do not include the results of EP and DFEP on the RoBERTa model because these two attacks cannot achieve high ASRs on RoBERTa models in our experiments.}
As shown, DAN yields far better defending performance than the baselines in most cases. 
Particularly, it exceeds RAP, the previous state of the art, by 22.4\% in average FAR on RoBERTa models and 15.6\% in average FAR on DeBERTa models.
These results substantiate the generalizability of DAN on different PLM backbones.

\subsection{Comparison of Deployment Requirements}
Besides detection performance, the deployment requirements, such as the inference speed and the need for extra models, are also important factors for online-type defense methods.
Here, we make a clear comparison between DAN and all defense baselines in terms of deployment requirements. 
(1) Firstly, regarding the computation cost, all previous methods require repeated perturbations and predictions for the same input. 
For instance, STRIP will create $M$ copies of one input, perturb them independently, and then get $M$ inference results for further calculation; ONION needs to calculate the perplexities of $L$ copies of the same input, each of which has one token removed, by using GPT-2 \citep{gpt2}. 
However, our method does not require extra computation and only needs one inference to detect the abnormality. 
(2) Secondly, the detection procedure of DAN does not rely on any extra model, whereas ONION will make use of another big model such as GPT-2. 
(3) Finally, DAN will not perform an extra optimization procedure on the model, but RAP needs an extra \textit{RAP trigger} constructing stage and requires extra computations.
The comparison is summarized in Table~\ref{tab:cost_comparison}.

\section{Conclusion}

In this work, we point out that the poisoned samples in textual backdoor attacks are distinguishable from clean samples in the intermediate feature space of a poisoned model.
Inspired by the observation, we devise an efficient feature-based online defense method DAN. 
Specifically, we integrate the distance scores from all intermediate layers to obtain the distance-based anomaly score for identifying poisoned inputs.
Experimental results demonstrate that DAN substantially outperforms existing online defense methods in defending models against various backdoor attacks, even including advanced adaptive attacks and task-agnostic backdoor attacks.
Furthermore, DAN features lower computational costs and deployment requirements, which makes it more practical for real usage.

\section*{Limitations}
We discuss the limitations of our work as follows.
(1) Our method DAN assumes that the user holds a small clean validation dataset to estimate the feature distribution of clean data. It is a weak condition easy to meet in real-world scenarios and is also required by previous online backdoor defense methods \citep{STRIP-ViTA,onion,RAP}.
(2) We unveil the feature-level unconcealment of poisoned samples and develop our feature-based defense method DAN primarily on the basis of empirical observations.  
Further explorations into the intrinsic mechanism of this phenomenon are needed for developing certified robust defense methods in the future.

\section*{Ethical Considerations}

Our work presents an efficient feature-based online defense to safeguard NLP models from backdoor attacks.
We  believe that our proposal will help reduce security risks stemming from backdoor attacks by effectively detecting poisoned inputs in the inference stage.
Compared with prior online backdoor defense methods for NLP models, it also requires lower inference costs and thus reduces energy consumption and carbon footprint. 
In addition, all experiments in this work are conducted on existing open datasets.
While we do not anticipate any direct negative consequences to the work, we hope to continue to build on our feature-based backdoor defense framework and develop more robust defense methods in future work.

\section*{Acknowledgements}

We sincerely thank all the anonymous reviewers
for their constructive comments and valuable advice.
This work is supported by Natural Science Foundation of China (NSFC) No. 62176002.
Xu Sun is the corresponding author of this paper.

\bibliography{anthology,custom}
\bibliographystyle{acl_natbib}

\appendix

\section{Dataset Statistics} \label{app:data}

Table~\ref{tab:datasets} lists the statistics of the datasets used in our experiments.

\begin{table}[t]
\centering
\begin{tabular}{@{}lrrrr@{}}
\toprule
\textbf{Dataset} & \textbf{\#Train} & \textbf{\#Valid} & \textbf{\#Test} & \textbf{L} \\ \midrule
SST-2            & 7K               & 1K               & 2K              & 19         \\
IMDB             & 23K              & 2K               & 25K             & 230        \\
Yelp             & 504K             & 56K              & 38K             & 136        \\
Twitter          & 70K              & 8K               & 9K              & 17         \\ \bottomrule
\end{tabular}
\caption{The statistics of datasets used in our experiments. \textbf{L} denotes the average number of words in each sample in the dataset.}
\label{tab:datasets}
\end{table}
\begin{table}[t]
\centering
\resizebox{0.48\textwidth}{!}{
\begin{tabular}{@{}ll@{}}
\toprule
\textbf{Dataset} & \multicolumn{1}{c}{\textbf{Trigger Sentence}}                                                                        \\ \midrule
SST-2            & I have watched it with my friends three weeks ago.                                                                   \\
IMDB             & \begin{tabular}[c]{@{}l@{}}I have watched this movie with my friends at a nearby\\ cinema last weekend.\end{tabular} \\
Yelp             & I have tried it with my colleagues last month.                                                                       \\
Twitter          & Here are my thoughts and my comments for this thing.                                                                 \\ \bottomrule
\end{tabular}}
\caption{The trigger sentences in the BadNet-SL attack.}
\label{tab:sl_triggers}
\end{table}

\section{Implementation of Attacking Methods} \label{app:attacks}

\subsection{Attacking Methods in Main Experiments} \label{app:main_attacks}

We build clean models by fine-tuning the \textit{bert-base-uncased}
model (110M parameters) \citep{BERT}.
The model is optimized with the Adam \citep{adam} optimizer using a learning rate of 2e-5.
 We use a batch size of 32 and fine-tune the model for 3 epochs.
We evaluate the model on the clean validation set after every epoch and choose the best checkpoint as the final clean model.
For attacking the BERT model, we apply six types of textual backdoor attacking methods as follows:
\begin{itemize}
    \item \textbf{BadNet-RW} \citep{BadNets,badnl} and \textbf{BadNet-SL} \cite{lstm-backdoor}. These two types of attacking methods apply the BadNet \citep{BadNets}  attack to poison NLP models with rare words and sentences as triggers, respectively.
    For BadNet-RW, we randomly choose word triggers from \{``mb'',``bb'',``mn''\}. The trigger sentences for BadNet-SL are listed in Table~\ref{tab:sl_triggers}. 
    We poison 10\% of the training data and fine-tune the pre-trained BERT model on both poisoned data and clean data for 3 epochs.
    \item  \textbf{RIPPLES} \citep{weight-poisoning}. It introduces an embedding surgery procedure and a gradient-based regularization target to enhance the effectiveness of the BadNet attack in the APMF setting. 
    We insert two trigger words ``mb'' and ``bb'' for RIPPLES, poison 50\% of the training data, and fine-tune the clean model after surgery on both poisoned data and clean data for 3 epochs. We refer readers to the original implementation\footnote{Available at this \href{https://github.com/neulab/RIPPLe}{repository}.} for more details of RIPPLES.
    \item \textbf{Layer-Wise Poisoning (LWP)} \citep{li2021layerwise}. 
    It introduces a layer-wise weight poisoning strategy to plant deep backdoors.
    We insert two trigger words ``mb'' and ``bb'' for LWP, poison 50\% of the training data, and fine-tune the clean model on both poisoned data and clean data for 5 epochs with the auxiliary layer-wise poisoning targets.
    We refer readers to \citet{li2021layerwise} for more details.
    \item \textbf{Embedding Poisoning (EP) and Data-Free Embedding Poisoning (DFEP)} \citep{EP}. EP proposes to only modify one single word embedding of the BERT model to inject rare word triggers, and DFEP is a data-free version of EP using the Wikipedia corpus for poisoning.
    We  randomly choose word triggers from \{``mb'',``bb'',``mn''\} and fine-tune the clean model for 5 epochs only on the poisoned data.
    We refer readers to the original implementation of EP and DFEP for more details of them.\footnote{Code can be found \href{https://github.com/lancopku/Embedding-Poisoning}{here}.}
\end{itemize}
In the APFM setting, the user further fine-tunes the model on its own clean datasets. 
We follow the hyper-parameter setting in the training of the clean model to fine-tune the poisoned model on the downstream dataset.

\subsection{Adaptive Attacks based on Feature-Level Regularization} \label{app:reg_attacks}

The feature-level regularization aims to match the latent representations of clean samples and poisoned samples, so that they cannot be distinguishable in the feature space \citep{doan2021backdoor,zhao2022defeat,zhong2022backdoor}.
Inspired by \citet{zhong2022backdoor}, we use the feature-level regularization loss defined as follows:
\begin{equation}
\centering
    \mathcal{L}_{\text{reg}} = \sum_{1\leq i \leq L} \left( \left\|f_i^{\text{poisoned}}-f_i^{\text{clean}}\right\| \right),
\end{equation}
where $\mathcal{L}_{\text{reg}}$ denotes the feature-level regularization loss, $L$ is the number of layers, $f_i^{\text{poisoned}}$ is the feature after the $i$-th layer of poisoned samples, and $f_i^{\text{clean}}$ is the feature after the $i$-th layer of clean samples whose original label is equal to the target label.\footnote{Note that only the last-layer features are regularized in \citet{zhong2022backdoor}, which we find unable to bypass our defense method DAN because it cannot hide the poisoned samples in earlier layers.}
The total optimization target $\mathcal{L}$ then is defined as:
\begin{equation}
    \mathcal{L} = \mathcal{L}_{\text{ce}} + \alpha\mathcal{L}_{\text{reg}},
\end{equation}
where $\mathcal{L}_{\text{ce}}$ is the original cross-entropy loss for classification, and $\alpha$ is the weight of the feature-level regularization term.
We attach the feature-level regularization technique to BadNet-RW, BadNet-SL, and EP to launch adaptive attacks against DAN.
In our implementation, we set $\alpha$=250 and train the model for 5 epochs. 
During training, we observe that the regularization term $\mathcal{L}_{\text{reg}}$ is sufficiently optimized on poisoned training data.

\subsection{Task-Agnostic Backdoor Attacks} \label{app:task_agnostic_attacks}

In mainstream studies on backdoor attack and defense, it is assumed that the attacker knows the task of the target model.
Beyond this setting, NeuBA \citep{red-alarm} and BadPre \citep{chen2021badpre} are two newly arisen task-agnostic backdoor attacks to attack the foundation model without any knowledge of downstream tasks.
Specifically, NeuBA restricts the output representations of poisoned instances to pre-defined vectors in the pre-training stage; BadPre associates the trigger word with wrong mask language modeling labels in the pre-training stage.
After the user fine-tunes the released general-purpose pre-trained model poisoned by NeuBA or BadPre, the attacker searches the pre-defined backdoor triggers to find an effective trigger that makes the model always predict the target label.
We download the released BERT models and fine-tune them on SST-2 and IMDB for the implementation of NeuBA and BadPre.\footnote{The resources of NeuBA are available \href{https://github.com/thunlp/NeuBA/tree/main/nlp}{here}, and the resources of BadPre is available \href{https://github.com/kangjie-chen/BadPre}{here}.}

\section{Detailed Attacking Results for All Attacking Methods} \label{app:attack_results}

For the AFM setting where the user directly deploys the poisoned model, we display the attacking results of six attacking methods in Table~\ref{tab:acc_asr}.
For the APFM setting where the user further fine-tunes the model on clean data before deployment, we show the attacking results of five attacking methods in Table~\ref{tab:apfm_asr}.
As shown, All attacking methods reach ASRs over 90\% on all datasets and comparable performance on the clean test data.
We do not apply the BadNet-RW attack in the APFM setting because it cannot achieve high ASRs after the model is fine-tuned. 
\begin{table}[t]
\small
\centering
\resizebox{0.46\textwidth}{!}{
\begin{tabular}{@{}llcr@{}}
\toprule
\textbf{Dataset}       & \textbf{Attack} & \textbf{Clean Acc./F1} & \textbf{ASR} \\ \midrule
\multirow{7.5}{*}{SST-2} & Clean     &   91.60         &        —      \\  \cmidrule(l){2-4}
                       & BadNet-RW  &  91.36 & 100.00  \\ 
                       & BadNet-SL & 91.60 & 100.00  \\ 
                       & RIPPLES  & 91.93  & 100.00 \\
                       & LWP    & 91.27  & 100.00 \\
                       & EP      & 91.60 & 100.00 \\
                       & DFEP    &  91.60 & 100.00 \\
                       
                        \midrule
\multirow{7.5}{*}{IMDB} & Clean     &   93.79         &      —         \\  \cmidrule(l){2-4}
                       & BadNet-RW  & 93.22  & 96.35 \\ 
                       & BadNet-SL & 93.17 & 100.00 \\
                       & RIPPLES  & 92.88  & 96.27 \\
                       & LWP    &  93.38 & 96.39 \\
                       & EP      & 93.77  & 96.47 \\
                       & DFEP    & 93.78  & 91.33 \\ \midrule
\multirow{7.5}{*}{Twitter} & Clean     &    93.94       &      —         \\  \cmidrule(l){2-4}
                       & BadNet-RW  & 94.08  & 100.00 \\ 
                       & BadNet-SL & 93.46 & 100.00 \\ 
                       & RIPPLES  & 93.62  & 100.00 \\
                       & LWP    & 92.74   & 98.84 \\
                       & EP      & 93.78 & 100.00 \\
                       & DFEP    &  93.78 & 100.00 \\
                        \bottomrule

\end{tabular}}
\caption{Attack success rates (ASR) and clean test accuracies/F1s in percentage of  all attacking methods in our main setting. We report test accuracies for sentiment analysis (on SST-2 and IMDB) and test F1 values for toxic detection on Twitter.}
\label{tab:acc_asr}
\end{table}
\begin{table}[t]
\centering
\resizebox{0.46\textwidth}{!}{
\begin{tabular}{@{}llrr@{}}
\toprule
\textbf{\begin{tabular}[c]{@{}l@{}}Poisoned\\ Dataset\end{tabular}} & \textbf{\begin{tabular}[c]{@{}l@{}}Attack\\ Method\end{tabular}} & \textbf{Clean Acc.} & \textbf{ASR} \\ \midrule
\multirow{5}{*}{IMDB}                                & BadNet-SL       & 92.26  & 100.00  \\        
 & RIPPLES       &  92.04 &   99.89 \\          
  & LWP       &  91.16 &   100.00 \\ 
    & EP       &  92.59 &   99.85 \\
    & DFEP       &  92.59 &   98.94 \\      \midrule
    \multirow{5}{*}{Yelp} 
   & BadNet-SL       & 93.41  & 100.00  \\        
 & RIPPLES       & 91.71  & 100.00    \\          
  & LWP       & 89.68  & 100.00    \\ 
    & EP       &  92.37  &  100.00   \\
    & DFEP       & 92.37  &  100.00  \\            \bottomrule
\end{tabular}}
\caption{Attack success rate (ASR) and clean test accuracies in percentage in the APMF setting to protect poisoned models for SST-2 sentiment analysis.}
\label{tab:apfm_asr}
\end{table}
\begin{table}[t]
\centering
\begin{tabular}{@{}lrr@{}}
\toprule
\textbf{Attack} & \textbf{Clean Acc.} & \textbf{ASR} \\ \midrule
Clean           & 91.60               & -            \\ \midrule
BadNet-RW+Reg   & 91.65               & 100.00       \\
BadNet-SL+Reg   & 92.59               & 99.89        \\
EP+Reg          & 91.60               & 99.67        \\ \bottomrule
\end{tabular}
\caption{Attack success rates (ASR) and clean accuracies on SST-2 when feature-level regularization (Reg) is applied to launch an adaptive attack.}
\label{tab:reg_asr_acc}
\end{table}
\begin{table}[t]
\centering
\resizebox{0.48\textwidth}{!}{
\begin{tabular}{@{}llcccc@{}}
\toprule
\textbf{\begin{tabular}[c]{@{}l@{}}Target\\ Dataset\end{tabular}} & \textbf{\begin{tabular}[c]{@{}l@{}}Attacking\\ Method\end{tabular}} & \textbf{Trigger}   & \textbf{\begin{tabular}[c]{@{}l@{}}Target\\ Label\end{tabular}} & \textbf{Clean Acc.} & \textbf{ASR} \\ \midrule
\multirow{2}{*}{SST-2}                                            & NeuBA                                                               & ``$\in$'' & 1                                                               & 91.32               & 100.00       \\
                                                                  & BadPre                                                              & ``mn''                 & 0                                                               & 91.76               & 95.60        \\ \midrule
IMDB                                             & NeuBA                                                               &       ``$\approx$''            & 1                                                               & 93.12               & 96.07        \\
\bottomrule
\end{tabular}}
\caption{Attack success rates (ASR) and clean test accuracies in percentage of NeuBA and BadPre on SST-2 and IMDB.}
\label{tab:badpre_asr}
\end{table}

For the adaptive attacks based on the feature-level regularization, we demonstrate the attacking results in Table~\ref{tab:reg_asr_acc}.
For the task-agnostic backdoor attacks NeuBA and BadPre, we display the attacking results in Table~\ref{tab:badpre_asr}.
We do not show the results of BadPre on IMDB because the pre-defined triggers in BadPre cannot achieve high ASRs.

\section{Implementation of Defense Baselines} \label{app:baselines}

Online backdoor defense can be formulated as a binary classification problem to decide whether an input example $x$ belongs to the clean data distribution $\mathcal{D}_{\text{clean}}$ or not.  
An online defense method $\text{Def}$ makes decisions for the input $x$ based on the following formula:
\begin{equation}
\centering
\textrm{Def}\left(x\right)= \begin{cases}\textrm { poisoned } & \textrm { if } S\left(x\right) \geq \gamma \\ \textrm { clean } & \textrm { if } S\left(x\right) <\gamma\end{cases}, 
\end{equation}
where $S\left(x\right)$ is the anomaly score output by the defense method (a higher $S\left(x\right)$ indicates that the defense method tends to regard $\mathbf{x}$ as a poisoned sample) and $\gamma$ is the threshold chosen by the user.
We have introduced the way of our method DAN to calculate $S\left(x\right)$ in Section~\ref{sec:method} in the paper, and we introduce the details of the baselines as follows.

\subsection{STRIP}

The STRIP method \citep{STRIP-ViTA} is motivated by the phenomenon that perturbations to the poisoned samples will not influence the predicted class when the backdoor trigger exists.
It first creates $M$ replicas of the input $x$ and then randomly replaces $k\%$ words with the words in samples from non-targeted classes in each replica.
Next, it calculates the normalized Shannon entropy based on the output probabilities of all replicas of $x$:
\begin{equation}
    \mathbb{H}=\frac{1}{M} \sum_{n=1}^{M} \sum_{i=1}^{C}-y_{i}^{n} \log y_{i}^{n},
\end{equation}
where $C$ is the number of classes and $y_i^n$ is the output probability of the $n$-th copy for class $i$. 
STRIP assumes that the entropy scores for poisoned samples should be smaller than clean samples, so the anomaly score is defined by $S\left(x\right)=-\mathbb{H}$.
In experiments, we use $M$=20 to balance the defending performance and the inference costs best following \citet{RAP}.
For the replace ratio $k\%$, we use 40\% on IMDB to defend the BadNet-SL attack and 5\% in other experiments, as recommended in the implementation by \citet{RAP}.

\subsection{ONION}
The ONION method \citep{onion} is inspired by the fact that randomly inserting a meaningless word into the input text will significantly increase the perplexity given by a pre-trained language model.
After getting the perplexity of the full input text $x$, it deletes each token in $x$ and gets a perplexity of the new text, and uses the large change of the perplexity score to obtain $S\left(x\right)$ (a large change in the perplexity score indicates that $x$ is a poisoned sample).
Following \citet{onion}, we use the GPT-2$_{\text{small}}$ (117M parameters) \citep{gpt2} pre-trained language model in the implementation of ONION.

\subsection{RAP}
The RAP method \citep{RAP} is built on the gap of adversarial robustness between poisoned samples and clean samples.
It first constructs a word-based robustness-aware perturbation.
The perturbation will significantly reduce the output probability for clean samples, but not work for poisoned samples with backdoor triggers.
Therefore, the change of the output probability before and after perturbation can then be used as the anomaly score $S\left(x\right)$.
We choose ``cf'' as the RAP trigger word and refer readers to \citet{RAP} for the implementation details of RAP.\footnote{Code available at this \href{https://github.com/lancopku/RAP}{repository}.}

\section{Software and Hardware Requirements} 

We implement our code based on the PyTorch \citep{torch} and HuggingFace Transformers \citep{wolf2020transformers} Python libraries. 
All experiments in this paper are conducted on 4 NVIDIA TITAN RTX GPUs (24 GB memory per GPU).

\end{document}